\documentclass[letterpaper]{article} 
\usepackage{aaai25}  
\usepackage{times}  
\usepackage{helvet}  
\usepackage{courier}  
\usepackage[hyphens]{url}  
\usepackage{graphicx} 
\usepackage{natbib}  
\usepackage{caption} 
\usepackage{algorithm}
\usepackage{algorithmic}
\usepackage{amsmath}
\usepackage{multirow}
\usepackage{comment}
\usepackage{graphicx}
\usepackage{makecell}
\usepackage{amssymb}
\usepackage{dsfont} 
\usepackage{booktabs}
\usepackage{xcolor}
\usepackage{newfloat}
\usepackage{listings}
\DeclareCaptionStyle{ruled}{labelfont=normalfont,labelsep=colon,strut=off} 
\floatstyle{ruled}
\newfloat{listing}{tb}{lst}{}
\floatname{listing}{Listing}
\title{Beyond Skip Connection: Pooling and Unpooling Design for Elimination Singularities}
\author{
    Chengkun Sun\textsuperscript{\rm 1},
    Jinqian Pan\textsuperscript{\rm 1},
    Zhuoli Jin\textsuperscript{\rm 3},
    Russell Stevens Terry\textsuperscript{\rm 2},
    Jiang Bian\textsuperscript{\rm 1},
    Jie Xu\textsuperscript{\rm 1}
}
\affiliations {
    \textsuperscript{\rm 1}Department of Health Outcomes and Biomedical Informatics, University of Florida, Gainesville, FL 32611, USA\\
    \textsuperscript{\rm 2}Department of Urology, University of Florida, Gainesville, FL 32611, USA\\
    \textsuperscript{\rm 3}Department of Statistics and Applied Probability, University of California
    Santa Barbara, CA 93106-3110, USA\\
    sun.chengkun@ufl.edu,
    jinqianpan@ufl.edu,
    zhuoli\_jin@pstat.ucsb.edu,
    russell.terry@urology.ufl.edu,
    bianjiang@ufl.edu,
    xujie@ufl.edu
}

\usepackage{bibentry}
\nocopyright
\begin{document}

\maketitle

\begin{abstract}
Training deep Convolutional Neural Networks (CNNs) presents unique challenges, including the pervasive issue of elimination singularities—consistent deactivation of nodes leading to degenerate manifolds within the loss landscape. These singularities impede efficient learning by disrupting feature propagation. To mitigate this, we introduce \textbf{Pool Skip}, an architectural enhancement that strategically combines a Max Pooling, a Max Unpooling, a $3\times3$ convolution, and a skip connection. This configuration helps stabilize the training process and maintain feature integrity across layers. We also propose the \textbf{Weight Inertia} hypothesis, which underpins the development of Pool Skip, providing theoretical insights into mitigating degradation caused by elimination singularities through dimensional and affine compensation. We evaluate our method on a variety of benchmarks, focusing on both 2D natural and 3D medical imaging applications, including tasks such as classification and segmentation. Our findings highlight Pool Skip's effectiveness in facilitating more robust CNN training and improving model performance.
\end{abstract}

%
\section{Introduction}
Convolutional Neural Networks (CNNs) are pivotal in advancing the field of deep learning, especially in image processing~\cite{jiao2019survey,razzak2018deep}. However, as these networks increase in depth to enhance learning capacity, they often encounter a notable degradation in performance~\cite{he2016deep}. This degradation manifests as a saturation point in accuracy improvements, followed by a decline, a phenomenon primarily driven by optimization challenges including vanishing gradients~\cite{he2016deep}. The introduction of Residual Networks (ResNets) with skip connections marked a significant advancement in mitigating these issues by preserving gradient flow during deep network training~\cite{he2016deep,orhan2017skip}. 

Despite these advancements, very deep networks, such as ResNets with upwards of 1,000 layers, still face pronounced degradation issues~\cite{he2016deep}. A critical aspect of this problem is the elimination singularity (ES)—stages in the training process where neurons consistently deactivate, producing zero outputs and creating ineffective nodes within the network~\cite{orhan2017skip,qiao2019rethinking}. This condition not only disrupts effective gradient flow but also significantly compromises the network's learning capability. ES often results from zero inputs or zero-weight configurations in convolution layers, which are frequently observed due to the tendency of training processes to drive weights towards zero, contributing to excessively sparse weight matrices~\cite{orhan2017skip,huang2020convolution}.
Additionally, the widely used Rectified Linear Unit (ReLU) activation function exacerbates these issues by zeroing out all negative inputs~\cite{qiao2019rethinking,lu2019dying}. This phenomenon, known as Dying ReLU, causes neurons to remain inactive across different data points, effectively silencing them and further complicating the training of deep networks~\cite{qiao2019rethinking,lu2019dying}.

To address these persistent challenges, we developed \textbf{Pool Skip}, a novel architectural module that strategically incorporates Max Pooling, Max Unpooling, and a $3 \times 3$ convolution linked by a skip connection. This design is specifically engineered to counteract elimination singularities by enhancing neuron activity and preserving the integrity of feature transmission across network layers. Our approach not only aims to stabilize the learning process but also to enhance the robustness of feature extraction and representation in deep networks. The key contributions of our work are summarized below:

\begin{itemize}
   \item We propose the \textbf{Weight Inertia} hypothesis to explain how persistent zero-weight conditions can induce network degradation. Based on this theory, we developed the Pool Skip module, which is positioned between convolutional layers and the ReLU function to help mitigate the risks associated with elimination singularities. We also provide mathematical proofs that demonstrate how Pool Skip's affine compensation and dimensional Compensation optimize gradient fluctuations during the backpropagation process, thus addressing the degradation problem at a fundamental level.

   \item We evaluated the proposed Pool Skip module across various deep learning models and datasets, including well-known natural image classification benchmarks (e.g., CIFAR-10 and CIFAR-100), segmentation tasks (e.g., Pascal VOC and Cityscapes), and medical imaging challenges (e.g., BTCV, AMOS). Our findings validate the effectiveness of Pool Skip in reducing elimination singularities and demonstrate its capacity to enhance both the generalization and performance of models.
\end{itemize}

\section{Related Work}
\subsection{Pooling Operations in CNNs}
Max Pooling~\cite{ranzato2007sparse}, a staple in CNN architectures, segments convolutional output into typically non-overlapping patches, outputting the maximum value from each to reduce feature map size~\cite{dumoulin2016guide,gholamalinezhad2020pooling}.
This not only yields robustness against local transformations but also leverages the benefits of sparse coding ~\cite{boureau2010theoretical, boureau2011ask, ranzato2007sparse}. Its efficacy is well-documented in prominent CNN architectures like VGG~\cite{simonyan2014very}, YOLO~\cite{redmon2016you}, and UNet~\cite{ronneberger2015u}, and is essential in numerous attention mechanisms, such as CBAM~\cite{woo2018cbam}, for highlighting salient regions.
Despite these advantages, Ruderman et al.~\cite{ruderman2018pooling} indicate that networks can maintain deformation stability without pooling, primarily through the smoothness of learned filters. Furthermore, Springenberg et al.~\cite{springenberg2014striving} suggest that convolutional strides could replace Max Pooling, as evidenced in architectures like nnUNet~\cite{isensee2021nnu}.


Max Unpooling, designed to reverse Max Pooling effects by restoring maximum values to their original locations and padding zeros elsewhere, complements this by allowing CNNs to learn mid and high-level features~\cite{zeiler2014visualizing, zeiler2011adaptive}.
However, the traditional ``encoder and decoder'' architecture, foundational to many modern CNNs like UNet~\cite{ronneberger2015u}, rarely adopts Max Unpooling due to concerns that zero filling can disrupt semantic consistency in smooth areas~\cite{liu2023learning}.

Our work reevaluates the conventional combination of Max Pooling and Max Unpooling, arguing that its effective utilization can still substantially benefit CNNs by focusing on significant features. Moreover, the finite number of layers in common encoder and decoder architectures limits the use of Max Pooling in deeply nested CNNs, posing challenges for deep-level information recognition.

\begin{figure*}[ht]
  \centering
  \includegraphics[width=0.85\textwidth]{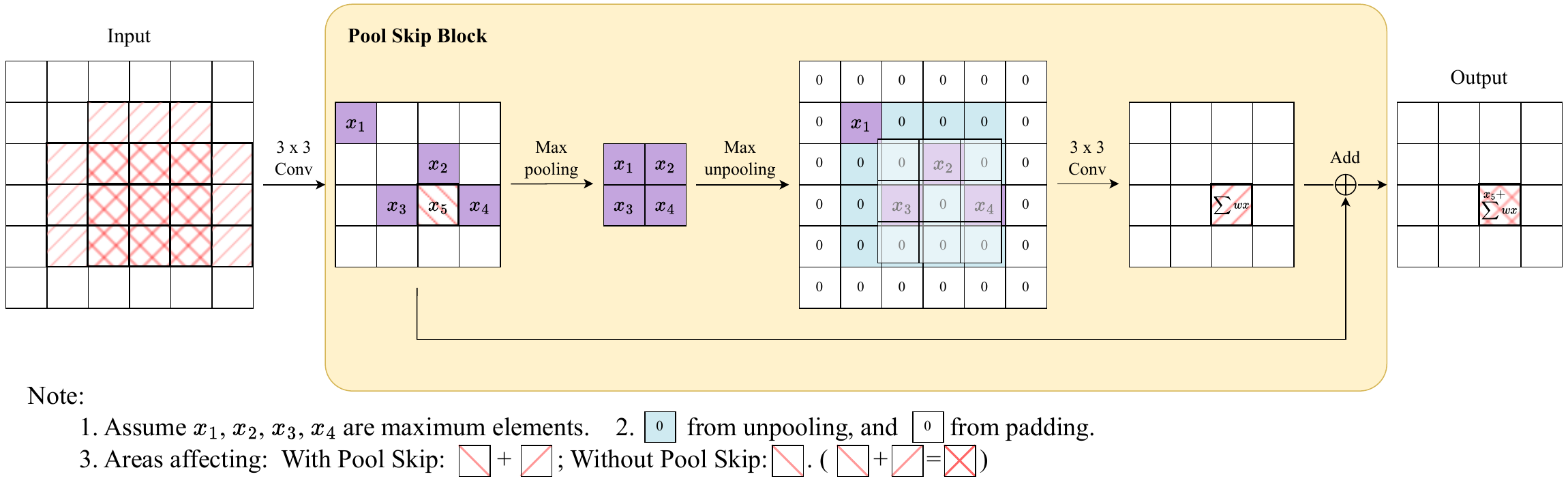}
  \caption{Schematic representation of the computational process of Pool Skip.}
  \label{pool_skip_calculation}
  \vspace{-0.3cm}
\end{figure*}

\subsection{Skip Connection and Batch Normalization}


The concept of ES was first proposed by Wei et al.~\cite{wei2008dynamics}, to describe the issue of zero weights in the output of convolutional layers, a phenomenon commonly referred to as ''weight vanishing''~\cite{wei2008dynamics}. This issue is particularly concerning because these zero weights do not contribute to the model's calculations, leading to inefficiencies in learning processes. ES is deeply associated with slow learning dynamics and unusual correlations between generalization and training errors and presents a significant challenge in training deep neural networks effectively~\cite{amari2006singularities}.

To mitigate ES, two primary strategies have emerged: normalization, specifically Batch Normalization (BN)~\cite{ioffe2015batch}), and skip connections~\cite{he2016deep, he2016identity}. BN helps maintain a stable distribution of activation values throughout training, while skip connections effectively increase network depth by preventing the elimination of singularities, ensuring that even with zero incoming or outgoing weights, certain layers maintain unit activation~\cite{ioffe2015batch, he2016deep}. This allows for the generation of non-zero features, making previously non-identifiable neurons identifiable, thereby addressing ES challenges~\cite{orhan2017skip}.
Despite these advancements, the degradation problem persists in extremely deep networks, even with the implementation of skip connections~\cite{he2016deep}. Further analysis by He et al.~\cite{he2016identity} on various ResNet-1001 components—such as constant scaling, exclusive gating, shortcut-only gating, conv shortcut, and dropout shortcut—reveals that the degradation issue in the original ResNet block not only remains but is also exacerbated.

\section{Pool Skip Mechanism}
In this section, we introduce the Pool Skip mechanism, beginning with the Weight Inertia hypothesis that motivates its development. We then provide theoretical insights into how this hypothesis helps mitigate degradation caused by elimination singularities through dimensional and affine compensation.

\subsection{Weight Inertia Hypothesis}
In the context of back-propagation~\cite{rumelhart1986learning}, the process is defined by several essential components. $L$ denotes the loss function, $W$ represents the weights, $Y$ refers to the output feature maps, $X$ to the input feature maps, while $c_{in}$ and $c_{out}$ indicate the input and output channels, respectively. The $\ast$ is used for the convolution operation. The operation of back-propagation is captured as follows: 
\begin{equation}
\footnotesize
\begin{aligned}
\frac{\partial L}{\partial W_{c_{in},c_{out}} } &= \frac{\partial L}{\partial Y_{c_{out}} }\times \frac{\partial Y_{c_{out}}}{\partial W_{c_{in},c_{out}} }\\
&=\frac{\partial L}{\partial Y_{c_{out}} }\times \frac{\partial \text{ReLU}(X_{c_{in}}\ast W_{c_{in},c_{out}}) }{\partial W_{c_{in},c_{out}} }\\
\frac{\partial L}{\partial X_{c_{in}} } &= \frac{\partial L}{\partial Y_{c_{out}} }\times \frac{\partial Y_{c_{out}} }{\partial X_{c_{in}} }\\&=\frac{\partial L}{\partial Y_{c_{out}} }\times \frac{\partial \text{ReLU}(X_{c_{in}}\ast W_{c_{in},c_{out}}) }{\partial X_{c_{in}} }
\end{aligned}
\label{back_propagation1}
\end{equation}
The activation function employed in this context is ReLU. During the convolution process, when the output $X_{c_{in}} \ast W_{c_{in},c_{out}}$ is less than or equal to zero, ReLU sets both derivatives, $\frac{\partial L}{\partial W_{c_{in},c_{out}}}$ and $\frac{\partial L}{\partial X_{c_{in}}}$ to zero, in accordance with its operational rules. Or if the weights themselves ($W_{c_{in},c_{out}}$) are zero, it would still result in zero gradients.


According to the standard gradient descent update rule, the weights are adjusted by subtracting a portion of the gradient from the current weight values: $\hat{W}_{c_{in},c_{out}} = W_{c_{in},c_{out}} - \eta \frac{\partial L}{\partial W_{c_{in},c_{out}} }$, where $\eta$ represents the learning rate. When training utilizes a fixed input space (a consistent set of training samples) and employs a diminishing learning rate towards the end of the training cycle, the updates to the weights in both the current and previous layers become minimal. This minimal update results in inputs $X_{c_{in}}$ and output $X_{c_{in}}\ast W_{c_{in},c_{out}}$ at the current layer becoming inert. Consequently, the outputs $Y_{c_{in}}$ of subsequent layers also remain unchanged. The worst case is a continuous negative or zero output.

This stagnation, which we term \textbf{Weight Inertia}, results from sparse weights (i.e. ES) and consistent non-positive outputs (i.e Dying ReLU), particularly prevalent in what we refer to as the degradation layer. This layer is marked by a continuous inability to update zero weights, leading to a limited number of effective weights and exacerbating the degradation problem. This forms a self-reinforcing cycle: as weights fail to update, the network's ability to learn and adapt diminishes, deepening the degradation. To break this cycle, controlling the negative outputs, specifically $X_{c_{in}} \ast W_{c_{in},c_{out}}$, is crucial. By effectively managing these outputs, it is possible to interrupt the cycle, prevent further degradation, and enhance the network's overall learning capabilities.

Motivated by the weight inertia hypothesis, we design Pool Skip, which consists of a Max Pooling layer, a Max Unpooling layer, and a $3 \times 3$ convolutional layer, tightly interconnected with skip connections spanning from the beginning to the end of the module. Figure~\ref{pool_skip_calculation} illustrates the computational process of Pool Skip. Initially, the Max Pooling layer prioritizes important features, facilitating the extraction of key information crucial for subsequent processing. Subsequently, the Max Unpooling layer ensures that the feature size remains consistent, preserving the gradient propagation process established by max-pooling. This characteristic allows Pool Skip to be seamlessly integrated into convolutional kernels at any position within the network architecture. Moreover, by selectively zeroing out non-key positions, Pool Skip effectively controls the magnitude of the compensatory effect, further enhancing its utility in stabilizing the training process.

\subsection{Affine and Dimensional Compensation}
\begin{figure}[h]
  \centering
  \includegraphics[width=0.45\textwidth]{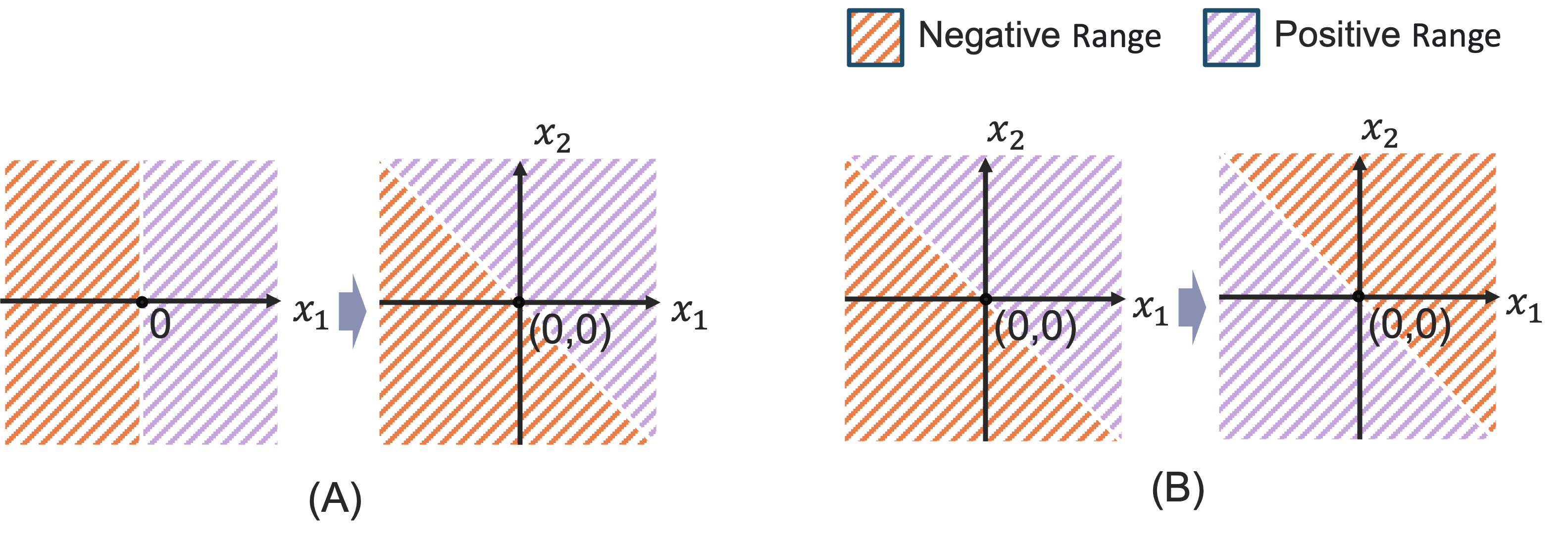}
  \vspace{-0.3cm}
  \caption{An simple example of dimensional and affine and compensation. 
  }
  \label{Compensation}
  \vspace{-0.2cm}
\end{figure}

As discussed earlier in the weight inertia hypothesis, the output of a neural network is often predominantly determined by a linear combination of a few specific and influential input elements, despite the potential presence of numerous input elements. This selective influence suggests that modifying this fixed linear combination—either by activating input elements through changes in input dimensions or by adjusting the coefficients of the linear combination—can significantly impact the output. For convenience, we refer to these two adjustment mechanisms as dimensional compensation and affine compensation. A simple example illustrating this process is provided in Figure~\ref{Compensation}. Initially, with one dimension $x_{1}$, the region where $x_{1}<0$ is shown in orange on the left side of Fig. (A). Introducing a second dimension $x_{2}$ changes this to $x_{1} + x_{2} <0$, shifting the orange area to the right side of Figure (A). This demonstrates dimensional compensation. When the coefficient in front of $x_{1} $ and $x_{2}$ changes from 1 to -1, the shift in the orange area in Fig. (B) represents affine compensation. These compensations alter the negative region of the input space, thereby disrupting Weight Inertia.

Next, we will theoretically explain how Pool Skip introduces dimensional and affine compensation, subsequently affecting the output results. 
As depicted in Figure \ref{pool_skip_calculation}, we begin by establishing the input configuration based on the computational process of Pool Skip:
\vspace{-0.3cm}
\begin{table}[htbp]
  \centering
    \begin{tabular}{|p{8cm}|}
    \hline
    1. $X_{H\times W}=\{x_{i,j}\}_{H\times W}$: input matrix;\\
    2. $W_{M\times M}=\{w_{i,j}\}_{M\times M}$: the convolutional kernel before Pool Skip. Assume $W$ is $M\times M$ kernel, and $M$ is an odd number; \\
    3. $Y_{(H-M+1) \times (W-M+1)}=\{y_{i,j}\}_{(H-M+1) \times (W-M+1)}$: the output of first convolutional computation;\\
    4. $e$: Max Pooling size which satisfies $e|H$, $e|W$, $e|H-M+1$ and $e|W-M+1$;\\
    5. $A_{c\times d}$: the matrix obtained from max-pooling on $Y$. $c=(H-M+1)/e$ and $d=(W-M+1)/e$;\\
    6. $\tilde{W}_{3 \times 3} = \{ \tilde{w}_{i,j} \}_{3 \times 3}$: the convolutional kernel in the Pool Skip;\\
    7. $O_{H-M+1, W-H+1} = \{o_{i, j}\}_{H-M+1, W-H+1}$: the output matrix.\\
    \hline
    \end{tabular}%
    \vspace{-0.3cm}
\end{table}%

\begin{table*}[htbp]
  \centering
    \begin{tabular}{|p{17cm}|}
    \hline
    \vspace{-0.6cm}
\begin{equation}
\begin{aligned}
\textbf{Final Output:} & \ \ o_{i, j} = y_{i, j} + y_{out, i, j}\\
&=
\left\{
\begin{aligned}
&\sum_{\mathds{1}_{K_i}((m,s))=1 \text{ and } \mathds{1}_{L_j}((n,t))=1}
[(1 + \tilde{w}_{s, t}) \times w_{m, n} \times x_{i - 1 + m, j - 1 + n}] + \\
&\sum_{\mathds{1}_{K_i}((m,s)) \neq 1 \text{ or } \mathds{1}_{L_j}((n,t)) \neq 1}\hspace{-15mm} (w_{m, n} \times x_{i-1+m, j-1+n} + \tilde{w}_{s, t} \times w_{m, n} \times x_{ue + \tilde{i_a}^{(u, v)} + m + s, ve + \tilde{j_a}^{(u, v)} + n + t}) \\
&
\quad\quad\quad\quad\quad\quad\quad\quad\text{if\ } e\ mod\ (i-1) = \tilde{i_a}^{(u, v)} \text{\ and\ } e\ mod\ (j-1) = \tilde{j_a}^{(u, v)} \text{\ in\ block\ }Y^{(u,v)}, \\
& \sum_{m=0}^{M-1}\sum_{n=0}^{M-1} w_{m, n} \times x_{i - 1 + m, j -1 + n}, 
\quad\quad\quad\quad\quad\quad\quad\quad\quad\quad\quad\quad\quad\quad\quad\quad\quad\quad\quad\quad\quad\text{o.w.}
\label{pool_skip_equation}
\end{aligned}\right.\\
&\text{for all $i\in \{1,2,\cdots, H-M+1\}$ and $j\in \{1,2,\cdots, W-M+1\}$.}
\end{aligned}
\end{equation}\\
\hline
    \end{tabular}%
    \vspace{-0.3cm}
\end{table*}%

In the convolutional computation of a single layer, the output $y_{i,j}$ is derived from a linear combination of the input $x_{i,j}$ before the Pool Skip~\cite{goodfellow2016deep}: 
\begin{equation}  
\begin{aligned}
y_{i, j} &= \sum_{m=0}^{M-1}\sum_{n=0}^{M-1} w_{m, n} \times x_{i + m, j + n}.
\end{aligned}
\label{Convolution}
\end{equation}
It's important to note that the convolution kernel is not flipped in this context. Based on the Max-Pooling, we can divide the $Y$ from previous computation by size $e\times e$ into $c\times d$ blocks. For each block of $Y$, indexed as $Y^{(u,v)}$, we have:
\begin{equation}
\begin{aligned}
    Y^{(u, v)} = 
\left(
  \begin{array}{ccc}
    y^{(ue, ve)} & \cdots & y^{(ue, (v+1)e-1)} \\
    \vdots & \ddots & \vdots \\
    y^{((u+1)e-1, ve)} & \cdots & y^{((u+1)e-1, (v+1)e-1)}\\
  \end{array}
\right)\\_{e \times e}, \\
\end{aligned}
\end{equation}
where $u \in \{ 0, 1, \cdots, c-1 \}, v \in \{ 0, 1, \cdots, d-1 \}$. For each block $Y^{(u,v)}$, the maximum element is 
\begin{equation}
\begin{aligned}
    \tilde{y}^{(u, v)} &= \mathop{\max}_{i^{(u, v)}, j^{(u, v)}\in\{0,1,\cdots,e-1\}} Y^{(u, v)}_{i^{(u, v)}, j^{(u, v)}},
\end{aligned}
\end{equation}
with the corresponding $\tilde{i_a}^{(u, v)}, \tilde{j_a}^{(u, v)}$ as
\begin{equation}
\begin{aligned}
    (\tilde{i_a}^{(u, v)}, \tilde{j_a}^{(u, v)}) &= \mathop{\arg\max}_{i^{(u, v)}, j^{(u, v)}\in\{0,1,\cdots,e-1\}} Y^{(u, v)}_{i^{(u, v)}, j^{(u, v)}}.
\end{aligned}
\end{equation}
Therefore, we can write $A$ as $A=\{\tilde{y}^{(u,v)}\}_{c\times d}$. And then via Max Unpooling and padding operations, we could get:

\begin{equation}
    y'_{i,j}=\left\{
    \begin{aligned}
        & \tilde{y}^{(u,v)}_{(\tilde{i_a}^{(u, v)},\tilde{j_a}^{(u, v)})}, 
            \text{if\ } e\ mod\ (i-1) = \tilde{i_a}^{(u, v)} \\
            &\text{\ and\ } e\ mod\ (j-1) = \tilde{j_a}^{(u, v)} \text{\ in\ block\ }Y^{(u,v)}\\
        & 0, \text{o.w.}\\
    \end{aligned}\right.
\end{equation}
where $1\leq i\leq H-M+1 \text{\ and\ } 1\leq j \leq W-M+1$. After passing the $3\times3$ convolutions, the result of the convolutional computation is shown as:
\begin{equation}
   y_{out, i, j} = \sum_{s=0}^{2}\sum_{t=0}^{2} \tilde{w}_{s, t} \times y'_{i + s, j + t}.
\end{equation}
After introducing the Pool Skip, the output $o_{i, j}$ is calculated by Eq.~(\ref{pool_skip_equation}), where we denote set:
\begin{equation}
\begin{aligned}
&K_i=\{(m,s):ue + \tilde{i_a}^{(u, v)} + m \\
&+ s \in ([i, i + M] \cap [ ue + \tilde{i_a}^{(u, v)}, ue + \tilde{i_a}^{(u, v)} + M + 2])\}, \\
&L_j=\{(n,t):ve + \tilde{j_a}^{(u, v)} + n \\ 
&+ t \in ([j, j+M] \cap [ve + \tilde{j_a}^{(u, v)},  ve + \tilde{j_a}^{(u, v)} + M + 2])\}.\\
\end{aligned}
\end{equation}
Note that when $(m,s)\in K_i$, $ue + \tilde{i_a}^{(u, v)} + m + s = i + m$, and when $(n,t)\in L_j$, $ve + \tilde{j_a}^{(u, v)} + n + t = j + n$.

According to Eq.~(\ref{pool_skip_equation}) (see detailed derivation in supplementary material), the original linear combination obtained two types of compensation. When the maximum value obtained after passing through the Pool Skip consists of the original linear combination elements $x$ (from the input feature of the convolutional kernel before the Pool Skip), part of the $x$ coefficients changed from $w_{m, n}$ to $(1 + \tilde{w}_{s, t}) \times w_{m, n}$, representing affine compensation. This change correlates closely with the weights of the convolutional kernel in Pool Skip. On the other hand, the remaining maximum values, which cannot be added to the original linear combination of $x$, expand the output dimensions 
(i.e. adding $\tilde{w}_{s, t} \times w_{m, n} \times x_{ue + \tilde{i_a}^{(u, v)} + m + s, ve + \tilde{j_a}^{(u, v)} + n + t}$), 
constituting dimensional compensation. This compensation also remains closely tied to the weights of the convolutional kernels in the Pool Skip.

These adjustments not only alter the contribution of input elements but also affect the negative range space of the output. This effectively breaks the constraints imposed by weight inertia, promoting diversity in output results and updating zero weights. Additionally, by adjusting the size of the Max Pooling and Max Unpooling kernels, we can control the number of maximum values, directly influencing the strength of the compensatory effects. Specifically, when the size of pooling kernels is 1, indicating only one convolution in the skip connection, every output element receives compensation. After receiving the compensatory effect, the original negative value range changes, allowing the original linear combination to output a non-zero effective value after ReLU. This activation enables neurons in the next layer to be activated during forward propagation and ensures that convolutional kernels with zero weights before ReLU receive gradient updates during backpropagation, thus alleviating the ES problem.

\section{Experiments}
We integrated the proposed Pool Skip into various deep networks and conducted comprehensive evaluations across common image tasks, 
including classification as well as natural image and medical image segmentation, utilizing diverse datasets for robust validation. All models were equipped with BN and ReLU or ReLU variations following the convolutions, ensuring a standardized architecture for comparison. Furthermore, all models were trained using a single NVIDIA A100 GPU with 80G of memory to maintain consistency in computational resources. 

\subsection{Image Classification}
\paragraph{Datasets}
For the classification task, we utilized the \textbf{CIFAR} datasets~\cite{krizhevsky2009learning}. CIFAR-10 comprises 60,000 color images categorized into 10 classes. 
 CIFAR-100 consists of 60,000 images divided into 100 classes, with each class containing 600 images. 
The images are colored and share the same dimensions of $32 \times 32$ pixels. 

\paragraph{Comparison Methods} 
We evaluated the effectiveness of the Pool Skip across various CNN architectures, including MobileNet~\cite{howard2017mobilenets}, GoogLeNet~\cite{szegedy2015going}, VGG16~\cite{simonyan2014very}, ResNet18~\cite{he2016deep}, and ResNet34~\cite{he2016deep}. 
To ensure robustness, each model was trained with 5 diverse seeds on the official training dataset, and the average and standard deviation of the Top-1 error from the final epoch were calculated on the official test dataset. Moreover, to assess the impact of the Pool Skip, it was implemented in each convolutional layer rather than solely in the first layer. All the training settings followed Devries and Taylor’s work~\cite{devries2017improved}.

\paragraph{Experimental Results}
Our experimental findings, detailed in Table~\ref{tab:cifar}, showcase the performance enhancements observed across the CIFAR10 and CIFAR100 datasets. Notably, we observed moderate improvements ranging from 0.5 to 5.44 on CIFAR100 and from 0.03 to 2.74 on CIFAR10 in networks with fewer layers. However, the magnitude of improvement varied depending on the architecture of the network. Of particular significance was the notable enhancement observed for MobileNet upon the integration of the Pool Skip.

\begin{table}[htbp]
  \centering
  \small
    \begin{tabular}{lll}
    \toprule
    \multicolumn{1}{c}{\multirow{2}[2]{*}{\textbf{Model}}} & \textbf{CIFAR100} & \textbf{CIFAR10} \\
    \cline{2-3}
          & Top-1 error (\%)& Top-1 error  (\%)\\
    \hline
    MobileNet
    & 33.75 $\pm$ 0.24 & 9.21 $\pm$ 0.19  \\
    \ \ \ \ +ours & 28.31 $\pm$ 0.23 \textcolor[rgb]{ 1,  0,  0}{-5.44}  & 6.47 $\pm$ 0.20 \textcolor[rgb]{ 1,  0,  0}{-2.74}\\
    \hline
    GoogleNet
    & 22.95 $\pm$ 0.24 & 5.35 $\pm$ 0.19 \\
    \ \ \ \ +ours & 22.36 $\pm$ 0.32 \textcolor[rgb]{ 1,  0,  0}{-0.59} & 5.19 $\pm$ 0.14 \textcolor[rgb]{ 1,  0,  0}{-0.16} \\
    \hline
    VGG16
    & 27.84 $\pm$ 0.38 & 6.24 $\pm$ 0.18  \\
    \ \ \ \ +ours & 27.23 $\pm$ 0.21 \textcolor[rgb]{ 1,  0,  0}{-0.61} & 5.90 $\pm$ 0.24 \textcolor[rgb]{ 1,  0,  0}{-0.34}  \\
    \hline
    ResNet18
    & 24.06 $\pm$ 0.18 & 5.17 $\pm$ 0.15\\
    \ \ \ \ +ours & 23.32 $\pm$ 0.14 \textcolor[rgb]{ 1,  0,  0}{-0.74}& 5.10 $\pm$ 0.14 \textcolor[rgb]{ 1,  0,  0}{-0.07} \\
    \hline
    ResNet34
    & 22.69 $\pm$ 0.18 & 4.89 $\pm$ 0.07 \\
    \ \ \ \ +ours & \textbf{22.19 $\pm$ 0.22 \textcolor[rgb]{ 1,  0,  0}{-0.50}} & \textbf{4.86 $\pm$ 0.06 \textcolor[rgb]{ 1,  0,  0}{-0.03}}\\
    \bottomrule
    \end{tabular}%
\caption{The Top-1 error rates (Mean $\pm$ Std) for image classification on CIFAR 100 and CIFAR 10 datasets.}
\label{tab:cifar}
\end{table}%

\subsection{Natural Image Segmentation}
\paragraph{Datasets}
For this task, we utilized \textbf{Cityscapes}~\cite{cordts2016cityscapes} and PASCAL Visual Object Classes (VOC) Challenge (\textbf{Pascal VOC})~\cite{everingham2010pascal} datasets. Cityscapes offers a comprehensive insight into complex urban street scenes, comprising a diverse collection of stereo video sequences captured across streets in 50 different cities. 
Pascal VOC provides publicly accessible images and annotations, along with standardized evaluation software. 
For segmentation tasks, each test image requires predicting the object class of each pixel, with ``background'' designated if the pixel does not belong to any of the twenty specified classes. 

\paragraph{Comparison Methods}  
We evaluated the effectiveness of the Pool Skip on DeepLabv3+ models~\cite{chen2017rethinking,chen2018encoder}, utilizing ResNet101~\cite{he2016deep} and MobileNet-v2~\cite{sandler2018mobilenetv2} backbones. The Pool Skip was exclusively employed in the convolution of the head block, the first convolution of the classifier, and all the atrous deconvolutions to validate its compatibility with atrous convolution. The models were trained using the official training data and default settings in~\cite{chen2017rethinking,chen2018encoder}, and the Intersection over Union (IoU) of the final-epoch model was recorded on the official validation data. Five seeds were selected to calculate the mean and standard deviation of the results.

\paragraph{Experimental Results} 
As illustrated in Table \ref{tab:deeplab}, our experiments demonstrated a modest improvement in mIoU ranging from 0.16\% to 0.53\% for DeepLabv3+ models, considering the incorporation of only five layers of the Pool Skip (one in the Deeplab head, three in the atrous deconvolutional layers, and one in the classifier). This indicates the compatibility of the Pool Skip with atrous deconvolutions.

\begin{table}[htbp]
  \centering
  \small
    \begin{tabular}{lll}
    \toprule
    \multicolumn{1}{c}{\multirow{2}[2]{*}{\textbf{Model}}} &\textbf{Cityscapes} & \textbf{Pascal VOC} \\
    \cmidrule(r){2-3}
          & \multicolumn{1}{l}{mIoU (\%)} & \multicolumn{1}{l}{mIoU (\%)} \\
    \hline
    DLP\_MobileNet
    & 71.72 $\pm$ 0.49 & 66.40 $\pm$ 0.37\\
    \ \ \ \ +ours & 71.96 $\pm$ 0.35 \textcolor[rgb]{ 1,  0,  0}{+0.24} & 66.93 $\pm$ 0.49 \textcolor[rgb]{ 1,  0,  0}{+0.53}   \\
    \hline
    DLP\_ResNet101
    & 75.59 $\pm$ 0.30 & 74.83 $\pm$ 0.56   \\
    \ \ \ \ +ours & 75.89 $\pm$ 0.10 \textcolor[rgb]{ 1,  0,  0}{+0.30}& 74.99 $\pm$ 0.23 \textcolor[rgb]{ 1,  0,  0}{+0.16} \\
    \bottomrule
    \end{tabular}%
  \label{tab:addlabel}%
    \caption{The results of mIoU (Mean $\pm$ Std) for natural image segmentation on Cityscapes and Pascal VOC datasets. ``DLP'' denotes ``DeepLabv3+''.}
    \label{tab:deeplab}%
    \vspace{-0.2cm}
\end{table}%

\subsection{Medical Image Segmentation}
\paragraph{Datasets}
We used abdominal multi-organ benchmarks for medical image segmentation,  i.e.,~\textbf{AMOS}~\cite{ji2022amos} and Multi-Atlas Labeling Beyond the Cranial Vault (\textbf{BTCV})~\cite{BTCV2015} datasets.
AMOS is a diverse clinical dataset offering 300 CT (Computed Tomography) scans and 60 MRI (Magnetic Resonance Imaging) scans with annotations. 
The public BTCV dataset consists of 30 abdominal CT scans sourced from patients with metastatic liver cancer or postoperative abdominal hernia. 

\paragraph{Comparison Methods} 
We evaluated the Pool Skip using nnUNet~\cite{isensee2021nnu} and V-Net~\cite{milletari2016v}. For nnUNet, our implementation closely follows the nnUNet framework~\cite{isensee2021nnu}, covering data preprocessing, augmentation, model training, and inference. Scans and labels were resampled to the same spacing as recommended by nnUNet. We excluded nnUNet's post-processing steps to focus on evaluating the model's core segmentation performance. For a fair comparison, when reproducing nnUNet, we retained its default configuration. For V-Net~\cite{milletari2016v}, we adopted the preprocessing settings consistent with nnUNet.

For the BTCV dataset, 12 scans were assigned to the test set, and 18 to the training and validation set. 
From AMOS, 360 scans(containing CTs and MRIs) were divided into 240 for training and validation and 120 for testing, with a training-to-validation ratio of 4:1. 
We performed 5-fold cross-validation on all models, averaging their softmax outputs across folds to determine voxel probabilities. Our evaluation is based on the Dice Score~\cite{milletari2016v}, Normalized Surface Dice (NSD)~\cite{nikolov2018deep}, and mIoU metrics. 

For the V-Net model, we only added the Pool Skip to the first two encoders due to the odd size of feature map outputs by the final encoder. As for the nnUNet model, we applied the Pool Skip in each convolutional layer on all encoders when training on the BTCV dataset. When training on the AMOS dataset, Pool Skips were employed on all encoders except for the final encoder.

\paragraph{Experimental Results}
As illustrated in Table~\ref{Medical_image_segmentation}, this Pool Skip architecture applies to networks for 3D medical imaging segmentation tasks. Enhancement on V-Net and nnUNet demonstrate the Pool Skip's efficacy for complex image segmentation tasks. 

\begin{table}[htbp]
  \centering
\small
    \begin{tabular}{llll}
    \toprule
    \multicolumn{1}{c}{\multirow{2}[2]{*}{\textbf{Model}}} & \multicolumn{3}{c}{\textbf{BTCV}} \\
    \cmidrule(r){2-4}
          & DICE (\%)  & NSD (\%)   & mIoU (\%)  \\
    \hline
    \multicolumn{1}{c}{V-Net
    } & 78.32 & 70.77  & 68.08  \\
    \multicolumn{1}{c}{\ \ \ \ +ours} & 79.70 \textcolor[rgb]{ 1,  0,  0}{+1.38} & 72.35\textcolor[rgb]{ 1,  0,  0}{+1.58}& 69.48 \textcolor[rgb]{ 1,  0,  0}{+1.40} \\
    \hline
    \multicolumn{1}{c}{nnUNet
    } & 81.52 & 76.10  &  72.14 \\
    \ \ \ \ +ours & 82.47 \textcolor[rgb]{ 1,  0,  0}{+0.95} & 77.00 \textcolor[rgb]{ 1,  0,  0}{+0.90}    & 73.08 \textcolor[rgb]{ 1,  0,  0}{+0.94}  \\
    \toprule
    \multicolumn{1}{c}{\multirow{1}[2]{*}{}} & \multicolumn{3}{c}{\textbf{AMOS}} \\
    \hline
    \multicolumn{1}{c}{V-Net
    } & 77.15 & 62.97 & 66.32 \\
    \multicolumn{1}{c}{\ \ \ \ +ours} & 80.02 \textcolor[rgb]{ 1,  0,  0}{+2.87} & 67.32 \textcolor[rgb]{ 1,  0,  0}{+4.35} & 69.81 \textcolor[rgb]{ 1,  0,  0}{+3.49} \\
    \hline
    \multicolumn{1}{c}{nnUNet
    } & 89.75 & 85.58 & 83.13 \\
    \ \ \ \ +ours & 89.78 \textcolor[rgb]{ 1,  0,  0}{+0.03} & 85.52 \textcolor[rgb]{ .267,  .447,  .769}{-0.06} &  83.13 \textcolor[rgb]{ 1,  0,  0}{+0} \\
    \bottomrule
    \end{tabular}%
  \caption{The results of DICE, NSD and mIoU for medical image segmentation on two datasets.}
    \label{Medical_image_segmentation}
    \vspace{-0.2cm}
\end{table}%

\begin{figure}[h]
  \centering
  \includegraphics[width=0.48\textwidth]{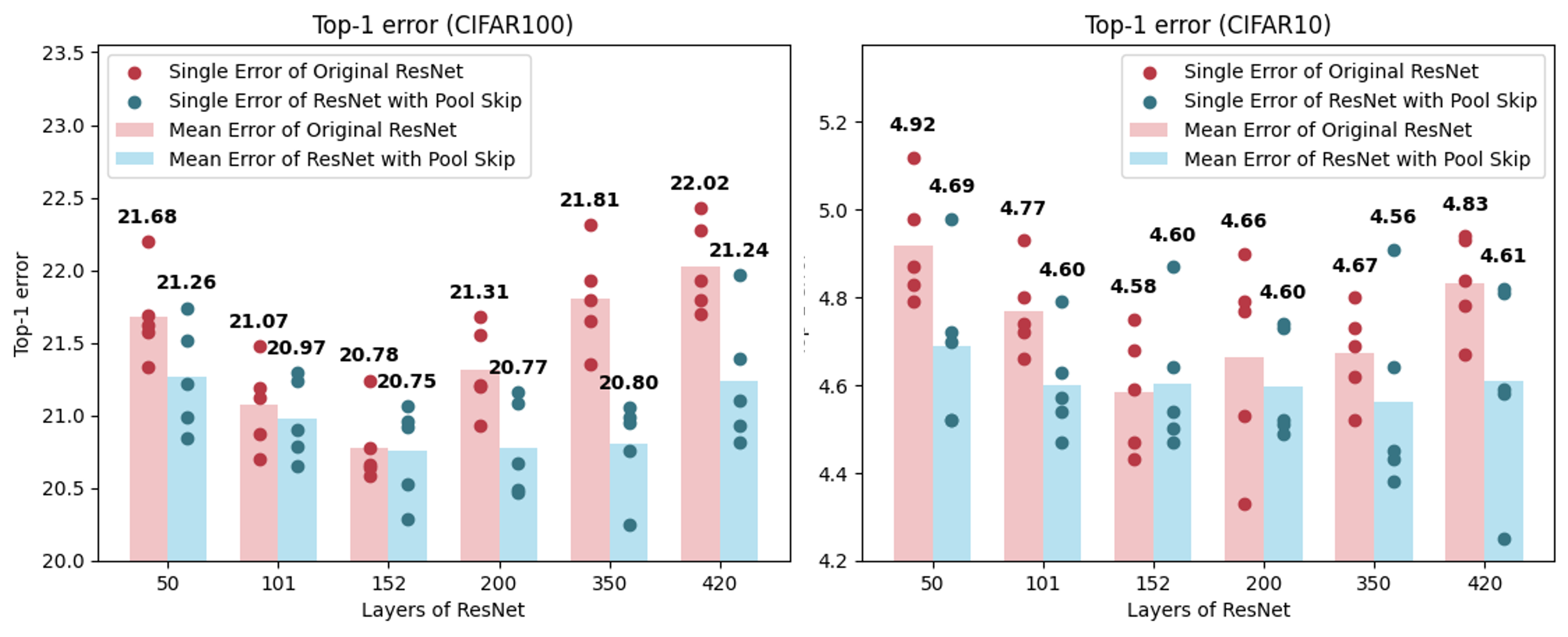}
  \caption{The Top-1 error rates of deep ResNet on CIFAR10 and CIFAR100 datasets. The pool kernel size is 4 for CIFAR100 experiments and 2 for CIFAR10 experiments.}
  \label{depth_curve}
\end{figure}

\subsection{Further Analysis}
\paragraph{Efficacy of Pool Skip in Deep CNNs}
To assess the effectiveness of the Pool Skip in deep CNNs, we utilized ResNet~\cite{he2016deep} as our baseline architecture. Our experiments covered a range of network depths, including 50, 101, 152, 200, 350, and 420 layers. For networks with fewer than 152 layers, we adhered to the original ResNet architecture as proposed by He et al. \cite{he2016deep}. For deeper architectures (i.e., 200, 350, and 420 layers), we followed the architectural specifications outlined in a prior study \cite{bello2021revisiting}. Training settings were consistent with those outlined in Devries and Taylor's work \cite{devries2017improved}.

The performance of the models across varying network depths is depicted in Figure~\ref{depth_curve}. Initially, as the number of layers increases, the model's performance improves before reaching a peak. The original ResNet achieves its best performance with 152 layers, achieving a top-1 error of 4.584 on CIFAR10 and 20.78 on CIFAR100. Subsequently, performance deteriorates rapidly. However, upon integrating the Pool Skip, performance improves, with a lower top-1 error of 4.562 on CIFAR10 with 350 layers and 20.752 on CIFAR100 with 152 layers. After stabilizing, performance begins to decline again. Nevertheless, there is a noticeable improvement in top-1 error (0.1-0.25 on CIFAR10 and 0.8-1 on CIFAR100) in deep networks. This underscores the efficacy of Pool Skip in enhancing the performance of deep CNNs.
Thus, Pool Skip demonstrates promise in mitigating the degradation phenomenon that both BN and ResNet struggle to address, particularly with the increase in model depth.

\begin{figure*}[htbp]
  \centering
  \includegraphics[width=0.9\textwidth]{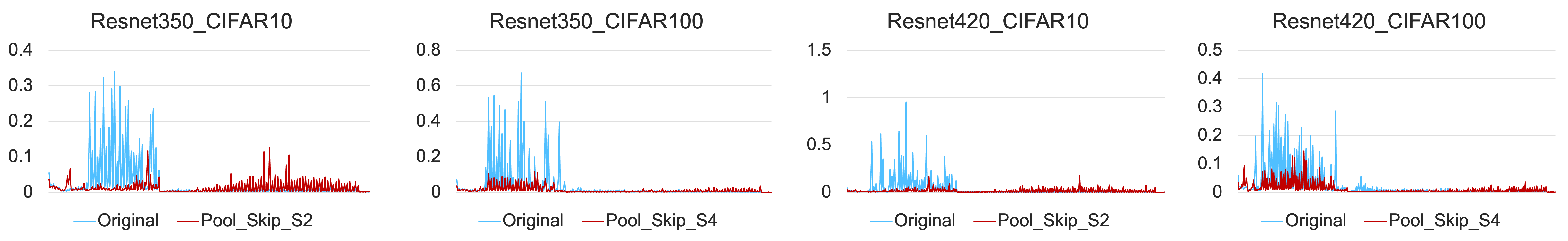}
  \caption{$\frac{l_2}{l_1}$ value quantitative comparison in ResNet350 and ResNet420 on CIFAR10 and CIFAR100 Datasets. The $\frac{l_2}{l_1}$ values were computed based on the output sequence of the network, with and without the incorporation of the Pool Skip. The plot highlights a moderate alleviation of the network degradation issue in shallow layers upon the integration of Pool Skip. Note: The horizontal axis represents the layers of the network along the output direction, from left to right. The ``Pool\_Skip\_S4'' means the size of Pool operation kernel is 4, ``Pool\_Skip\_S4'' does 2.}
  \label{l2_l1}
  \vspace{-0.2cm}
\end{figure*}

\paragraph{Efficacy of Mitigating Elimination Singularities (ES)}
To assess weight sparsity, we calculated the $\frac{l_2}{l_1}$ ratio for each layer with and without the Pool Skip, as proposed by Hoyer et al.~\cite{hoyer2004non}. The higher the $\frac{l_2}{l_1}$ value is, the more zero the weights may contain~\cite{wang2020accelerated,hoyer2004non}. Figure~\ref{l2_l1} illustrates the $\frac{l_2}{l_1}$ curve.
Original ResNet350 and ResNet420 architectures suffered from severe degradation issues in shallow layers, despite the inclusion of batch normalization. However, integrating the Pool Skip noticeably alleviated this problem. Specifically, the $\frac{l_2}{l_1}$ ratios for ResNet350 decreased from a maximum of 0.7 to 0.1 on CIFAR100 and from 0.35 to 0.1 on CIFAR10. Similarly, for ResNet420, the ratios decreased from a maximum of 0.4 to 0.15 on CIFAR100 and from 1 to 0.15 on CIFAR10. This underscores the efficacy of the Pool Skip in mitigating elimination singularities, thereby enhancing model stability and maintaining feature integrity across layers.

\paragraph{Ablation Experiments}
We conducted ablation experiments based on VGG16, as detailed in Table \ref{ablation},  to evaluate the impact of each block on the overall Pool Skip. Skip connections were identified as the most crucial component of the Pool Skip, evidenced by a 13\% reduction in the Top-1 error upon their removal. This improvement in model performance cannot be solely attributed to an increase in network parameters. Notably, a decrease of 3\% in the Top-1 error was observed when only convolutions and skip connections were utilized. However, removing convolutions did not result in a significant change in model performance.

\begin{table}[htbp]
  \centering
  \small
    \begin{tabular}{ccccc}
    \toprule
    \multicolumn{1}{p{0.7cm}}{VGG16
    } & \multicolumn{1}{p{1.6cm}}{+{\text{\{Pool, Skip\}}}} & \multicolumn{1}{p{1.6cm}}{+{\text{\{Conv, Skip\}}}} & \multicolumn{1}{p{1.6cm}}{+{\text{\{Pool, Conv\}}}} & \multicolumn{1}{p{1.1cm}}{+{\text{\{Ours\}}}} \\
    \hline
    27.84 & 27.88 & 30.82 & 44.07 & \textcolor[rgb]{ 1,  0,  0}{27.23} \\
    \bottomrule
    \end{tabular}%
\caption{The Top-1 error rates (\%) of ablation experiments on VGG16: ``Pool'' denotes Max Pooling and Max Unpooling, ``Conv'' represents $3\times3$ convolutions, and ``Skip'' indicates skip connections.}
\label{ablation}
\vspace{-0.3cm}
\end{table}%

\section{Discussion}
While Pool Skip holds promise for mitigating the elimination singularities issue in convolutional kernels and has demonstrated effectiveness across extensive datasets and network architectures, the proposed structure still has some limitations. Firstly, since each convolutional kernel is followed by a $3\times3$ convolution, the overall number of parameters in the network increases significantly, thereby adding a burden to both training and inference processes. Additionally, the sizes of Max Pooling kernels used in the experiments with deep ResNet are 2 and 4, while in compatibility experiments, it is 2. However, this structure cannot be applied when the size of feature maps is not multiples of the chosen sizes, especially during the encoder phase when downsampling reduces the feature maps to an odd size. Additionally, the effectiveness of dimensional and affine compensations needs to be optimized through the adjustments of the pooling size in the Pool Skip.

\begin{table}[h]
  \centering
  \small
    \begin{tabular}{lll}
    \toprule
    \multicolumn{1}{l}{\multirow{2}[2]{*}{\textbf{Model}}} & \textbf{CIFAR100} & \textbf{CIFAR10} \\
    \cline{2-3}
          & Top-1 error (\%)& Top-1 error  (\%)\\
    \hline
    ViT
    & 25.37 $\pm$0.17  &  6.85 $\pm$0.23 \\
    +ours & 25.07 $\pm$0.33  \textcolor[rgb]{ 1,  0,  0}{-0.30}  & 7.07 $\pm$0.27  \textcolor[rgb]{ 1,  0,  0}{+0.22}\\
    \hline
    CCT
    & 19.16 $\pm$0.20 & 3.87 $\pm$0.19\\
    +ours & 18.61 $\pm$0.12 \textcolor[rgb]{ 1,  0,  0}{-0.55} & 3.89 $\pm$0.14  \textcolor[rgb]{ 1,  0,  0}{+0.02} \\
    \hline
    CVT
    & 22.92 $\pm$0.38 & 5.98 $\pm$0.19  \\
    +ours & 22.80 $\pm$0.40 \textcolor[rgb]{ 1,  0,  0}{-0.12} & 6.36 $\pm$0.36  \textcolor[rgb]{ 1,  0,  0}{+0.38}  \\
    \bottomrule
    \end{tabular}%
\caption{Top1-error on Vit-based model}
\label{vit}
\vspace{-0.3cm}
\end{table}%

On the other hand, ViTs generally rely on one or more convolutional layers to generate image patches. To explore potential improvements in performance, we experimented with integrating the pool skip directly into the patch generation process. Our goal was to observe how this modification influences the overall effectiveness of the ViTs. We evaluated three ViT-based models including original ViT~\cite{dosovitskiy2020image}, CCT~\cite{hassani2021escaping} and CVT~\cite{wu2021cvt} on the CIFAR10 and CIFAR100 datasets. According to Table~\ref{vit}, pool skip gains from 0.12 to 0.55 reduction in Top-1 error on CIFAR100, but from 0.02 to 0.38 deterioration. We believe that the observed performance changes are likely due to the pooling layer's ability to extract key features, leading to performance improvements. However, the potential information loss caused by pooling may also contribute to performance deterioration. Given that this model was initially proposed to address the elimination of singularities in shallow networks, there remains much to explore regarding its application within Vision Transformers (ViTs). This also represents a limitation in the application of this structure.

\section{Conclusion}
In this paper, we introduced the Pool Skip, a novel and simple architectural enhancement designed to mitigate the issue of elimination singularities in training deep CNNs. Our theoretical analysis, rooted in the Weight Inertia hypothesis, highlights how Pool Skip effectively provides affine and dimension compensatory effects, thereby stabilizing the training process. Through extensive experimentation on diverse datasets and models, we have demonstrated the efficacy of Pool Skip in optimizing deep CNNs and enhancing the learning capacity of convolutions.

\bibliography{aaai25}
\end{document}


\onecolumn
\section{Technical Appendix}
\subsection{Mathematical Derivation of Pool Skip}
\label{details_of_pool_skip}
\noindent  \textbf{The input setting:}
\begin{enumerate}
    \item $X_{H\times W}=\{x_{i,j}\}_{H\times W}$: input matrix.
    \item $W_{M\times M}=\{w_{i,j}\}_{M\times M}$: the convolutional kernel before Pool Skip. Assume $W$ is $M\times M$ kernel, and $M$ is an odd number. 
    \item $Y_{(H-M+1) \times (W-M+1)}=\{y_{i,j}\}_{(H-M+1) \times (W-M+1)}$: the output of first convolutional computation.
    \item $e$: max-pooling size which satisfies $e|H$, $e|W$, $e|H-M+1$ and $e|W-M+1$.
    \item $A_{c\times d}$: the matrix obtained from max-pooling on $Y$. $c=(H-M+1)/e$ and $d=(W-M+1)/e$.
    \item $\tilde{W}_{3 \times 3} = \{ \tilde{w}_{i,j} \}_{3 \times 3}$: the convolutional kernel in the Pool Skip.
    \item $O_{H-M+1, W-H+1} = \{o_{i, j}\}_{H-M+1, W-H+1}$: the output matrix.
\end{enumerate}

\vspace{0.5cm}\hrule \vspace{0.5cm}
\noindent \textbf{Step 1:} Apply $W_1$ to $X$ to obtain $Y$.

\begin{equation}
   y_{i, j} = \sum_{m=0}^{M-1}\sum_{n=0}^{M-1} w_{m, n} \times x_{i + m, j + n}
\end{equation}
Note now $Y$ is a $(H-M+1)\times (W-M+1)$ matrix.

\vspace{0.5cm}\hrule \vspace{0.5cm}
\noindent \textbf{Step 2:} Size $e$ max-pooling.

We can write $Y$ into $c\times d$ blocks, for each block we have
\begin{equation}
\begin{aligned}
    Y^{(u, v)} = 
\left(
  \begin{array}{ccc}
    y^{(ue, ve)} & \cdots & y^{(ue, (v+1)e-1)} \\
    \vdots & \ddots & \vdots \\
    y^{((u+1)e-1, ve)} & \cdots & y^{((u+1)e-1, (v+1)e-1)}\\
  \end{array}
\right)_{e \times e} \\
\end{aligned}
\end{equation}
where $u \in \{ 0, 1, \cdots, c-1 \}, v \in \{ 0, 1, \cdots, d-1 \}$. For each block $Y^{(u,v)}$, the maximum element is 
\begin{equation}
\label{eq::11}
\begin{aligned}
    \tilde{y}^{(u, v)} &= \mathop{\max}_{i^{(u, v)}, j^{(u, v)}\in\{0,1,\cdots,e-1\}} Y^{(u, v)}_{i^{(u, v)}, j^{(u, v)}},
\end{aligned}
\end{equation}
with the corresponding $\tilde{i_a}^{(u, v)}, \tilde{j_a}^{(u, v)}$ as
\begin{equation}
\begin{aligned}
    (\tilde{i_a}^{(u, v)}, \tilde{j_a}^{(u, v)}) &= \mathop{\arg\max}_{i^{(u, v)}, j^{(u, v)}\in\{0,1,\cdots,e-1\}} Y^{(u, v)}_{i^{(u, v)}, j^{(u, v)}}.
\end{aligned}
\end{equation}
Therefore, we can write $A$ as $A=\{\tilde{y}^{(u,v)}\}_{c\times d}$.

\vspace{0.5cm}\hrule \vspace{0.5cm}

\noindent \textbf{Step 3:} Size $e$ max-unpooling.

First, we unpack $A$ to the $(H-M+1)\times (W-M+1)$ matrix $\hat{Y}$ such that the $c\times d$ blocks of $\hat{Y}$ be defined as 
\begin{equation}
\begin{aligned}
    \hat{Y}^{(u,v)} = \{\hat{y}^{(u,v)}_{i_a,j_a}\}_{e\times e},\quad \hat{y}_{i_a,j_a}^{(u,v)} =  Y_{i_a,j_a}^{(u,v)}\mathds{1}(\{(i_a,j_a) = (\tilde{i_a},\tilde{j_a})^{(u,v)}\})\\
    ,u\in \{0,1,\cdots, c-1\}, v\in \{0,1,\cdots, d-1\}.
\end{aligned}
\end{equation}
\textbf{Transfer local indices $(\tilde{i_a},\tilde{j_a})^{(u,v)}$ back to global indices$(\tilde{i},\tilde{j})^{(u,v)}$}, we have 
\begin{equation}
\label{eq::14}
    \hat{y}_{i,j} = \left\{
\begin{aligned}
    & \tilde{y}^{(u,v)}_{(\tilde{i_a}^{(u,v)},\tilde{j_a}^{(u,v)})}, \quad \text{if\ }e\ mod\ i = \tilde{i_a}^{(u,v)}\text{\ and\ }e\ mod\ j = \tilde{j_a}^{(u,v)}\text{\ in\ block\ }(u,v) \\
    &0, \quad \text{o.w.}
\end{aligned}
\right.
\end{equation}

where $u = \left\lfloor \frac{i}{e} \right\rfloor$, $v =  \left\lfloor \frac{j}{e} \right\rfloor$, $\tilde{i_a} = i\ mod\ e$ and $\tilde{j_a} = j\ mod\ e$. 

Add extra rows and columns of 0 outside $\hat{Y}$, we have a $(H-M+3)\times (W-M+3)$ matrix $Y'=\{y'_{i,j}\}$ and each $y'_{i,j}$ can be represented as
\begin{equation}
    \label{eq::15}
    y'_{i,j}=\left\{
    \begin{aligned}
        & \hat{y}_{i+1,j+1}\\
        & 0, \quad \text{o.w.}
    \end{aligned}\right.
\end{equation}

where $1\leq i\leq H-M+1\text{\ and\ } 1\leq j \leq W-M+1$. 

Combine the Equation \ref{eq::15} and \ref{eq::14}:
\begin{equation}
\label{eq::16}
    y'_{i,j}=\left\{
    \begin{aligned}
        & \tilde{y}^{(u,v)}_{(\tilde{i_a},\tilde{j_a})}, 
            &\quad \text{if\ } e\ mod\ (i-1) = \tilde{i_a}^{(u,v)} \text{\ and\ } e\ mod\ (j-1) = \tilde{j_a}^{(u,v)} \text{\ in\ block\ }Y^{(u,v)} \\
        & 0, \quad &\text{o.w.}
    \end{aligned}\right.
\end{equation}
where $1\leq i\leq H-M+1\text{\ and\ } 1\leq j \leq W-M+1$. 

Combine the Equation \ref{eq::16} and \ref{eq::11}:
\begin{equation}
\label{eq::17}
    y'_{i,j}=\left\{
    \begin{aligned}
        & \mathop{\max}_{i_a, j_a} Y^{(u, v)}_{i_a, j_a}, 
            &\quad \text{if\ } e\ mod\ (i-1) = \tilde{i_a} \text{\ and\ } e\ mod\ (j-1) = \tilde{j_a} \text{\ in\ block\ }(u,v)\\
        & 0, \quad &\text{o.w.}
    \end{aligned}\right.
\end{equation}
where $1\leq i\leq H-M+1\text{\ and\ } 1\leq j \leq W-M+1$. 

\vspace{0.5cm}\hrule \vspace{0.5cm}

\noindent \textbf{Step 4:} $3 \times 3$ Conv.

\begin{equation}
\label{eq::18}
   y_{out, i, j} = \sum_{s=0}^{2}\sum_{t=0}^{2} \tilde{w}_{s, t} \times y'_{i + s, j + t}
\end{equation}
Note now $Y'$ is a $(H-M+1)\times (W-M+1)$ matrix. \\

\noindent Combine the Equation \ref{eq::18} and \ref{eq::17}:

\begin{equation}
    y_{out, i, j} = \left\{
    \begin{aligned}
         & \sum_{s=0}^{2}\sum_{t=0}^{2} \tilde{w}_{s, t} \times \mathop{\max}_{i_a, j_a} Y^{(u, v)}_{i_a, j_a}, \\
        &\quad\quad\quad \text{if\ } e\ mod\ (i-1-s) = \tilde{i_a} \text{\ and\ } e\ mod\ (j-1-t) = \tilde{j_a} \text{\ in\ block\ }(u,v)\\
        & 0, \quad\quad\quad\quad\quad\quad\quad\quad\quad\quad\quad\quad\quad\quad\quad\quad\quad\quad\quad\quad\quad\quad\quad\quad\quad\quad\quad\quad\quad \text{o.w.}
    \end{aligned}\right.
\end{equation}
where $1\leq i\leq H-M+1\text{\ and\ } 1\leq j \leq W-M+1$. 

\vspace{0.5cm}\hrule \vspace{0.5cm}
\noindent \textbf{Step 5:} Add Pool skip.

\begin{equation}
    o_{i, j} = y_{i, j} + y_{out, i, j}
\end{equation}

\noindent If $e\ mod\ (i-1-s) = \tilde{i_a}^{(u,v)}$ and $e\ mod\ (j-1-t) = \tilde{j_a}^{(u,v)}$ in block $(u,v)$, where $1\leq i\leq H-M+1$ and $1\leq j \leq W-M+1$.

\begin{equation}
\begin{aligned}
&o_{i, j} 
    = \sum_{m=0}^{M-1}\sum_{n=0}^{M-1} w_{m, n} \times x_{i - 1 + m, j - 1 + n} \\
    &+ 
    \sum_{s=0}^{2}\sum_{t=0}^{2} \tilde{w}_{s, t} \times
        \mathop{\max} \{ 
            \sum_{m=0}^{M-1}\sum_{n=0}^{M-1} w_{m, n} \times x_{ue + m +s, ve + n + t}, \\
            & \sum_{m=0}^{M-1}\sum_{n=0}^{M-1} w_{m, n} \times x_{ue + 1 + m + s, ve + 1 + n + t},
            \cdots,
            \sum_{m=0}^{M-1}\sum_{n=0}^{M-1} w_{m, n} \times x_{(u+1)e + m + s, (v+1)e + n + t}
            \} \\
    =&\sum_{m=0}^{M-1}\sum_{n=0}^{M-1} w_{m, n} \times x_{i- 1 + m, j- 1 + n} + 
\\
&\sum_{s=0}^{2}\sum_{t=0}^{2} \tilde{w}_{s, t} \times \sum_{m=0}^{M-1}\sum_{n=0}^{M-1} w_{m, n} \times x_{ue + \tilde{i_a}^{(u,v)} + m + s, ve + \tilde{j_a}^{(u,v)} + n + t}
\end{aligned}
\end{equation}

where $u \in \{ 0, 1, \cdots, c-1 \}, v \in \{ 0, 1, \cdots, d-1 \}.$\\

\noindent Denote set 
$K_i=\{(m,s):ue + \tilde{i_a}^{(u,v)} + m + s \in ([i, i + M] \cap [ ue + \tilde{i_a}^{(u,v)}, ue + \tilde{i_a}^{(u,v)} + M + 2])\}$ 
and $L_j=\{(n,t):ve + \tilde{j_a}^{(u,v)} + n + t \in ([j, j+M] \cap [ve + \tilde{j_a}^{(u,v)},  ve + \tilde{j_a}^{(u,v)} + M + 2])\}$. \\
Note that when $(m,s)\in K_i$, $ue + \tilde{i_a}^{(u,v)} + m + s = i- 1 + m$, and when $(n,t)\in L_j$, $ve + \tilde{j_a}^{(u,v)} + n + t = j- 1 + n$.

we have:

\begin{equation}
\begin{aligned}
&o_{i, j} 
= \sum_{m=0}^{M-1}\sum_{n=0}^{M-1} w_{m, n} \times x_{i- 1 + m, j- 1 + n} \\
&+ 
\sum_{s=0}^{2}\sum_{t=0}^{2} \tilde{w}_{s, t} \times \sum_{m=0}^{M-1}\sum_{n=0}^{M-1} w_{m, n} \times x_{ue + \tilde{i_a}^{(u,v)} + m + s, ve + \tilde{j_a}^{(u,v)} + n + t} \\
&= \sum_{\mathds{1}_{K_i}((m,s))=1 \text{ and } \mathds{1}_{L_j}((n,t))=1} \hspace{-10mm}
(w_{m, n} \times x_{ue + \tilde{i_a}^{(u,v)} + m + s, ve + \tilde{j_a}^{(u,v)} + n +t} + \tilde{w}_{s, t} \times w_{m, n} \\ &\times x_{ue + \tilde{i_a}^{(u,v)} + m + s, ve + \tilde{j_a}^{(u,v)} + n + t}) 
\\ &+ \sum_{\mathds{1}_{K_i}((m,s)) \neq 1 \text{ or } \mathds{1}_{L_j}((n,t)) \neq 1} \hspace{-5mm}(w_{m, n} \times x_{i+m, j+n} + \tilde{w}_{s, t} \times w_{m, n} \times x_{ue + \tilde{i_a}^{(u,v)} + m + s, ve + \tilde{j_a}^{(u,v)} + n + t}) \\
&= \sum_{\mathds{1}_{K_i}((m,s))=1 \text{ and } \mathds{1}_{L_j}((n,t))=1} \hspace{-5mm}
[(1 + \tilde{w}_{s, t}) \times w_{m, n} \times x_{ue + \tilde{i_a}^{(u,v)} + m + s, ve + \tilde{j_a}^{(u,v)} + n +t}] \\
&+ \sum_{\mathds{1}_{K_i}((m,s)) \neq 1 \text{ or } \mathds{1}_{L_j}((n,t)) \neq 1} \hspace{-5mm}(w_{m, n} \times x_{i+m, j+n} + \tilde{w}_{s, t} \times w_{m, n} \times x_{ue + \tilde{i_a}^{(u,v)} + m + s, ve + \tilde{j_a}^{(u,v)} + n + t}) \\
&= \sum_{\mathds{1}_{K_i}((m,s))=1 \text{ and } \mathds{1}_{L_j}((n,t))=1} \hspace{-5mm}
[(1 + \tilde{w}_{s, t}) \times w_{m, n} \times x_{i- 1 + m, j- 1 + n}] \\
&+ \sum_{\mathds{1}_{K_i}((m,s)) \neq 1 \text{ or } \mathds{1}_{L_j}((n,t)) \neq 1}\hspace{-10mm} (w_{m, n} \times x_{i- 1+m, j- 1+n} + \tilde{w}_{s, t} \times w_{m, n} \times x_{ue + \tilde{i_a}^{(u,v)} + m + s, ve + \tilde{j_a}^{(u,v)} + n + t})
\end{aligned}
\end{equation}

\vspace{0.5cm}\hrule \vspace{0.5cm}
Therefore:

\noindent Denote set 
$K_i=\{(m,s):ue + \tilde{i_a}^{(u,v)} + m + s \in ([i, i + M] \cap [ ue + \tilde{i_a}^{(u,v)}, ue + \tilde{i_a}^{(u,v)} + M + 2])\}$ 
and $L_j=\{(n,t):ve + \tilde{j_a}^{(u,v)} + n + t \in ([j, j+M] \cap [ve + \tilde{j_a}^{(u,v)},  ve + \tilde{j_a}^{(u,v)} + M + 2])\}$. \\
Note that when $(m,s)\in K_i$, $ue + \tilde{i_a}^{(u,v)} + m + s = i + m$, and when $(n,t)\in L_j$, $ve + \tilde{j_a}^{(u,v)} + n + t = j + n$.

\begin{equation}
\begin{aligned}
&o_{i, j} = y_{i, j} + y_{out, i, j}\\
&=
\left\{
\begin{aligned}
&\sum_{\mathds{1}_{K_i}((m,s))=1 \text{ and } \mathds{1}_{L_j}((n,t))=1}
[(1 + \tilde{w}_{s, t}) \times w_{m, n} \times x_{i- 1 + m, j- 1 + n}] \\
&+ \hspace{-5mm}\sum_{\mathds{1}_{K_i}((m,s)) \neq 1 \text{ or } \mathds{1}_{L_j}((n,t)) \neq 1}\hspace{-15mm} (w_{m, n} \times x_{i- 1+m, j- 1+n} + \tilde{w}_{s, t} \times w_{m, n} \times x_{ue + \tilde{i_a}^{(u, v)} + m + s, ve + \tilde{j_a}^{(u, v)} + n + t}) \\
&\quad\quad\quad\quad\quad \quad \quad\quad
\text{if\ } e\ mod\ (i-1) = \tilde{i_a}^{(u, v)} \text{\ and\ } e\ mod\ (j-1) = \tilde{j_a}^{(u, v)} \text{\ in\ block\ }Y^{(u,v)}, \\
& \sum_{m=0}^{M-1}\sum_{n=0}^{M-1} w_{m, n} \times x_{i- 1 + m, j- 1 + n}, \quad\quad\quad\quad\quad\quad \quad \quad\quad\quad\quad\quad\quad\quad\quad\quad\quad\quad\quad\quad \text{o.w.}
\end{aligned}\right.
\end{aligned}
\end{equation}
for all $i\in \{1,2,\cdots, H-M+1\}$ and $j\in \{1,2,\cdots, W-M+1\}$.

\clearpage